\documentclass[11pt]{article}

\usepackage[preprint]{acl}
\usepackage{times}
\usepackage{latexsym}
\usepackage[T1]{fontenc}
\usepackage[utf8]{inputenc}
\usepackage{microtype}
\usepackage{inconsolata}
\usepackage[most]{tcolorbox}
\usepackage{amsmath}
\usepackage{tabularx}
\usepackage{graphicx}
\usepackage{enumitem}
\usepackage{amssymb}
\usepackage{booktabs}
\usepackage{placeins} 
\usepackage{float}
\usepackage{multirow}
\usepackage{array}
\usepackage{tikz}
\usepackage{xcolor}
\usepackage{soul}
\usepackage{subcaption}
\usepackage{pgf-pie}
\usepackage{setspace}
\usetikzlibrary{positioning, shapes, arrows.meta}

\usepackage{multirow}
\usepackage{graphicx}
\usepackage[table,xcdraw]{xcolor}

\usepackage[most]{tcolorbox}

\usepackage{verbatim}
\usepackage{stfloats}
\usepackage{listings}

\newcounter{promptctr}
\renewcommand{\thepromptctr}{Prompt~\Alph{promptctr}}

\definecolor{examplebg}{HTML}{D1FFBD}
\definecolor{examplebg2}{HTML}{F1F1EE}

\usepackage{enumitem}
\usepackage{mdframed}

\newtcolorbox{exampleblock}{
  enhanced,
  colback=examplebg2,
  colframe=examplebg2,   
  boxrule=0pt,
  arc=2mm,
  outer arc=2mm,
  left=8pt,
  right=8pt,
  top=6pt,
  bottom=6pt,
  boxsep=0pt,
  before skip=8pt,
  after skip=8pt,
  sharp corners=all,
  breakable
}

\tcbset{
  bw:domain/.style={
    colback=white,
    colframe=black!40,
    coltitle=black,
    fonttitle=\bfseries\sffamily,   
    colbacktitle=examplebg,
    boxrule=0.5pt,
    titlerule=0pt,
    arc=2pt,
    left=4pt,right=4pt,top=4pt,bottom=4pt,
    toptitle=2pt,bottomtitle=2pt,
  }
}

\newcommand{\mytcbinputwide}[5]{
  \begin{figure*}[t]
  \centering
  \refstepcounter{promptctr}
  \phantomsection
  \begin{tcolorbox}[title={\thepromptctr: #2},#4,width=\textwidth,enhanced]
    \lstinputlisting{#1}
    \label{#5}
  \end{tcolorbox}
  \vspace{-4pt}
  \end{figure*}
}

\newcommand{\promptref}[1]{%
  \hyperref[#1]{\ref*{#1}}%
}

\newcommand{\promptrefp}[1]{%
  \hyperref[#1]{\ref*{#1} (p.~\pageref*{#1})}%
}

\newcommand{\system}{\textsc{AIG-Academic-Integrity-Guard}}

\newcommand{\igllm}{\textsc{AIG-LLM}}
\newcommand{\benchmark}{\textsc{Integrity-Bench}}
\newcommand{\fewsort}{\textsc{Few-SoRT}}
\newcommand{\fewsortq}{\textsc{FewSoRT-Q}}
\newcommand{\fewsortd}{\textsc{FewSoRT-D}}
\newcommand{\method}{\textsc{DoPE}}
\newcommand{\codeglyph}{\texttt{code-glyph}}
\newcommand{\trapdoc}{\textsc{TrapDoc}}
\title{\method{}: Decoy Oriented Perturbation Encapsulation \\ Human-Readable, AI-Hostile Documents for Academic Integrity}

\author{
Ashish Raj Shekhar\thanks{contributed equally} \quad
Shiven Agarwal\footnotemark[1] \quad
Priyanuj Bordoloi \quad
Yash Shah \\
\textbf{Tejas Anvekar} \quad
\textbf{Vivek Gupta} \\
Arizona State University \\
\texttt{\{ashekha9, sagar147, pbordolo, yshah124, tanvekar, vgupt140\}@asu.edu}
}

\begin{document}

\maketitle

\begin{abstract}
Multimodal Large Language Models (MLLMs) can directly consume exam documents, threatening conventional assessments and academic integrity. We present \method{} (Decoy-Oriented Perturbation Encapsulation), a document-layer defense framework that embeds semantic decoys into PDF/HTML assessments to exploit render–parse discrepancies in MLLM pipelines. By instrumenting exams at authoring time, \method{} provides model-agnostic prevention (stop or confound automated solving) and detection (flag blind AI reliance) without relying on conventional one-shot classifiers. We formalize prevention and detection tasks, and introduce \fewsortq{}, an LLM-guided pipeline that generates question-level semantic decoys and \fewsortd{} to encapsulate them into watermarked documents. We evaluate on \benchmark{}, a novel benchmark of 1826 exams (PDF+HTML) derived from public QA datasets and OpenCourseWare. Against black-box MLLMs from OpenAI and Anthropic, \method{} yields strong empirical gains: a 91.4\% detection rate at an 8.7\% false-positive rate using an LLM-as-Judge verifier, and prevents successful completion or induces decoy-aligned failures in 96.3\% of attempts. We release \benchmark{}, our toolkit, and evaluation code to enable reproducible study of document-layer defenses for academic integrity.
\end{abstract}

\begin{figure}[t]
  \centering
  \includegraphics[width=0.95\linewidth]{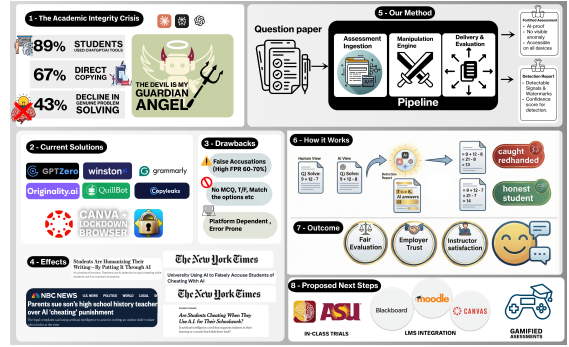}
  \caption{Illustration of integrates layered document protections with LMS delivery and verifiable detection.}
\end{figure}

\section{Introduction}
\label{sec:intro}


The release of ChatGPT in November 2022 marked a significant shift in the validity of educational assessments and academic integrity \cite{chatgpt2022,Sunjak2024ChatGPTTEA}. This led to the rapid adoption of AI-generated text detectors in educational settings to counter academic integrity violations\cite{Bao2024FastDetectGPT,pangram2024,pmlr-v202-mitchell23a}.
However, recent work has shown that these current detection approaches are not reliable at all due to several \cite{niu2024detecting}. Post-hoc text classifiers based on perplexity \citep{pmlr-v202-mitchell23a} or stylometric features \citep{pangram2024} suffer from systematic biases and flaws: \citet{liang2023gpt} found a 61.3\% false positive rate on TOEFL essays written by non-native English speakers, with 19.8\% unanimously misclassified as AI-generated by all seven tested detectors. These tools are trivially evaded through paraphrasing, with commercial humanizers achieving more than 90\% bypass rates \citep{sadasivan2023}. They also don't generalize to all kinds of Assessment items, such as MCQ, True/False, Match the Following, etc, as the lexical surface is insufficient for any kind of stylometric analysis.

Now keeping these inefficiencies of one-shot classifiers in mind, We propose a paradigm shift from \emph{passive detection: \textbf{Post-Hoc} detectors} to \emph{active instrumentation: \textbf{Pre-Hoc} instrumentation} on the assessment delivery methods,  similar to  \textbf{Watermarking} for peer-review journals such as works \cite{Liu2025InContextWatermarks,jin2025trapdoc}. We propose \method{}, a framework designed keeping in mind  the structure and constraints of academic assessments, that is applicable in Modern LMS Systems \cite{garcia2021canvasadoptionassessmentacceptance}, prominent across academic institutions.
Rather than analyzing student text post hoc, \method{} exploits the document processing capabilities of MLLMs \cite{keuper2025promptinjectionattacksllm, xiong2025invisible}: PDFs and HTML contain structural layers that render identically for humans but yield different content when parsed by MLLMs based on the document type and their internal configuration. By embedding imperceptible perturbations at assessment authoring time, we create documents that induce deterministic, detectable signals and errors in MLLM outputs while preserving human readability and functionality.
We make the following contributions:

\begin{enumerate}
  \item We introduce \method{} (Decoy-Oriented Perturbation Encapsulation), a PDF/HTML instrumentation that embeds semantic decoys exploiting render–parse gaps to \emph{prevent} and \emph{detect} blind MLLM assistance without model access or post-hoc text-only analysis.
  \item We propose, \fewsortq{} generates question-level semantic decoys; \fewsortd{} embeds them into documents. Together they yield a visually unchanged, \emph{shielded} exam that reliably induces decoy-aligned MLLM behavior.
  \item We release \benchmark{}, a paired corpus of 1{,}826 exams (PDFs+HTMLs) with multiple watermarked variants per exam for controlled evaluation of document-layer defenses.
  \item On black-box MLLMs (OpenAI, Anthropic) \method{} achieves strong prevention and detection, e.g., 91.4\% detection at 8.7\% FPR, supported by human imperceptibility checks and judge validation.
\end{enumerate}

\section{Related Work}
\label{sec:related}

\paragraph{The Academic Integrity Crisis.} Empirical evidence documents widespread AI adoption in academic contexts. The HEPI/Kortext Student Survey 2025 found 88\% of UK undergraduates use generative AI for assessments, up from 53\% in 2024 \citep{hepi2025}. Turnitin's analysis of over 250 million submissions identified 81\% containing at least partially AI-written content \citep{turnitin2024}. This correlates with documented cognitive effects: \citet{gerlich2025} established a significant negative correlation ($r = -0.68$) between frequent AI tool usage and critical thinking abilities, mediated by cognitive offloading.

\paragraph{Commercial Detectors and their Limitations.} Independent evaluations reveal substantial gaps between claimed and actual performance. GPTZero claims 99\% accuracy with about 380k reported users marked as instructors, but achieves 80\% in peer-reviewed evaluation with 10\% false positive rates \citep{liang2023gpt}. Vanderbilt University disabled Turnitin's AI detection in August 2023, noting that even 1\% false positives applied to 75,000 papers yields $\sim$750 wrongful accusations annually \citep{vanderbilt2023}. These studies also demonstrated systematic bias against non-native English speakers, with seven detectors showing 61.3\% false positive rates on TOEFL essays compared to near-zero on native English writing. This occurs because perplexity-based detection systematically penalizes simpler vocabulary and grammatical structures commonly used by English language learners, mistaking linguistic variation for evidence of AI generation. Compounding this issue, several commercial detector providers simultaneously offer AI \emph{``humanizer”} tools, undermining the reliability and ethical legitimacy of their own detection claims.

\paragraph{Text Adversarial Attacks.} Adversarial NLP traditionally perturbs question text to flip model outputs while preserving semantic meaning \citep{salim2024impedingllmassistedcheatingintroductory, ness2024medfuzz}. These approaches build on adversarially robust generalization, studying how intentional perturbations cause models to produce incorrect classifications or generated artifacts \citep{zou2023universal,Chao2023JailbreakingBB}. However, improvements in pre-processing and training have made frontier models increasingly robust to adversarial distractors and typographical noise. As a result, purely token-level edits either become perceptible to human readers or are normalized by the model pipeline, rendering them largely ineffective in real assessment settings.

\paragraph{Document-Layer Attacks.}
As MLLMs process documents, images, and other forms of input, they open a new vulnerability surface: by embedding hidden commands as white-colored text, font manipulation, or similar imperceptible alterations, PDFs and HTML webpages can be used to manipulate AI output \cite{xiong2025invisible,jin2025trapdoc, Liu2025InContextWatermarks}. We adopt this threat model as our ideal attack setting, focusing on perturbations that remain visually imperceptible to humans while being reliably parsed by document-processing pipelines of MLLMs. This enables adversarial control without modifying the model or relying on post-hoc detection, aligning with realistic deployment scenarios.

\paragraph{Watermarking Approaches.}
Text watermarking schemes \citep{pmlr-v202-kirchenbauer23a} bias token generation toward detectable patterns but require model access and are vulnerable to paraphrasing attacks. \citet{sadasivan2023} demonstrated that recursive paraphrasing breaks all tested schemes.
Document-layer watermarking offers an orthogonal approach that operates on the assessment rather than the response, thereby avoiding these limitations of not having access to model gradients.

\section{\method{}: Decoy-Oriented Perturbation Encapsulation}
\label{sec:dope}
\begin{figure*}[t]
  \centering
  \includegraphics[
    width=\linewidth]{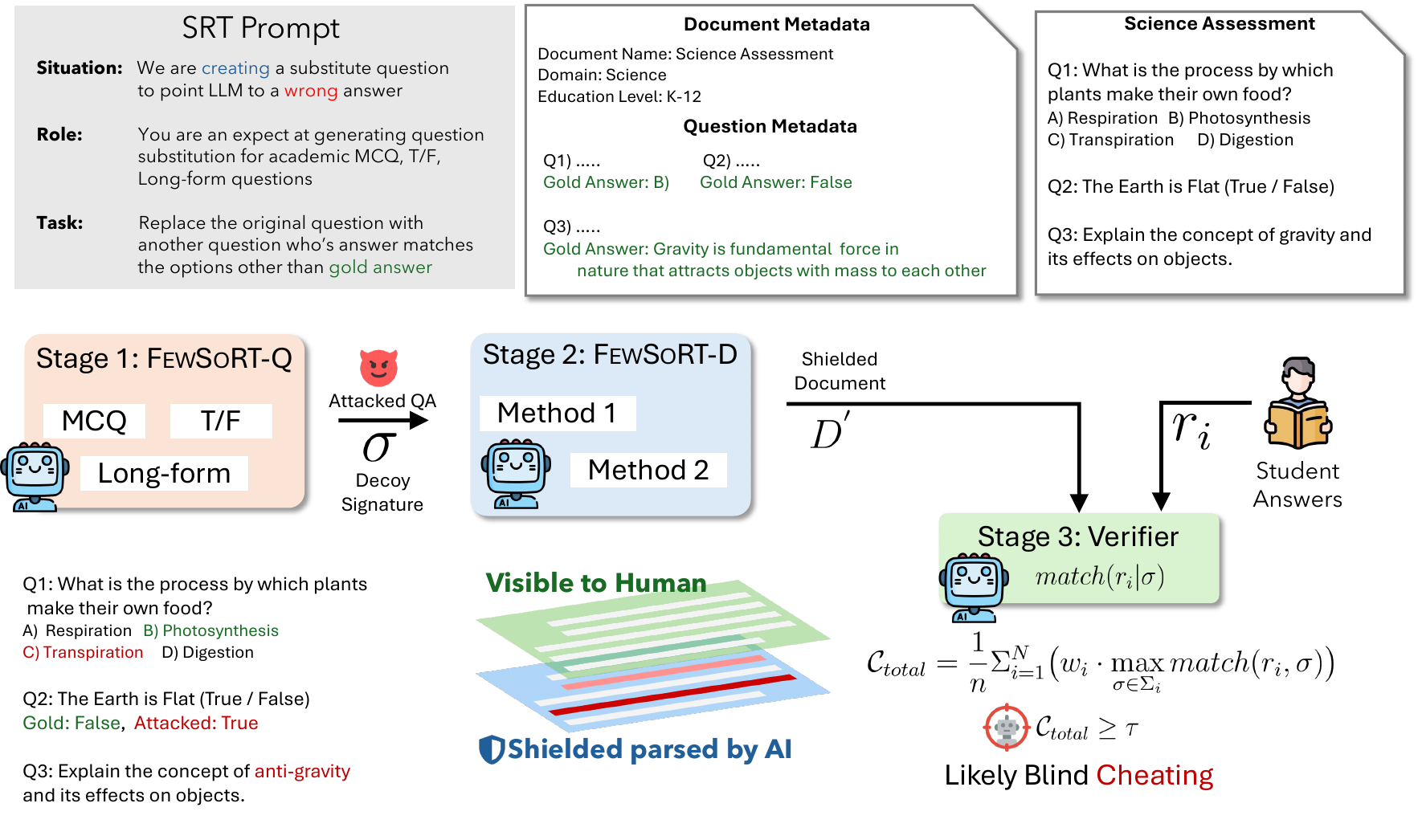}
  \caption{Overview of \method{}. The pipeline shows how
  \method{} creates AI-resistant assessments and detects misuse through document-layer perturbations.}
  \label{fig:integrity-shield-arch}
\end{figure*}

\method{} builds assessment documents that are fully readable and pedagogically correct for humans, yet systematically hostile to AI chatbots. The core idea is to embed \emph{decoy perturbations} that leave the human-visible rendering unchanged while altering the semantics parsed by Multimodal LLMs (MLLMs). Blind copying of AI outputs then produces predictable, detectable errors, whereas independent or critical use of AI rarely does.

\paragraph{Setting and assumptions.}
We consider a standard written assessment setting with a mix of item types. Multiple-choice and true/false questions must be accompanied by short written justifications, and long-form questions require free-text explanations. Students complete exams digitally and may either upload responses as PDFs/HTML or copy--paste their text into MLLMs to obtain assistance. We assume typical student behavior rather than specialized adversarial expertise. Instructors retain the official solutions and all perturbation metadata used to generate the watermarked exam documents.

\paragraph{Academic Integrity Guard.}
\method{} models behaviour along an \emph{Academic Integrity Guard} spectrum rather than a binary honest/cheating label. At one end are students who solve problems independently; at the other are students who submit MLLM outputs with minimal checking, with critical AI users in between. \method{} is calibrated so that blind reliance aligns strongly with decoy signatures, while genuine or edited reasoning remains largely unaligned.

\paragraph{Threat Models.}
We consider two workflows: (a.) In the stronger threat, a student uploads the entire assessment to an MLLM, which then processes hidden text, remapped fonts, and overlays and thus sees all decoys. (b.) In the weaker threat, the student copies only visible text; some perturbations survive, but coverage is reduced. \method{} is optimised for document upload, with partial robustness to copy-paste.

\paragraph{System Architecture.}
The \method{} pipeline has three components, illustrated in \autoref{fig:integrity-shield-arch}. \fewsortq{} generates semantic perturbations at the question level, \fewsortd{} embeds these perturbations into the document while preserving visual appearance, and the \emph{Verifier} scores student submissions against the induced decoy signatures. 

\subsection{\fewsortq{}: Question-Level Perturbation Generation}
\fewsortq{} is an LLM agent that constructs semantic perturbations with
\emph{predictable failure modes}. Unlike ICW-style control text or TrapDoc
local edits~\cite{Liu2025InContextWatermarks,jin2025trapdoc}, it reasons
explicitly over question semantics and answer distributions.

\fewsortq{} is driven by a compact \textbf{SRT} (\emph{Situation}, \emph{Role},
\emph{Task}) instruction that specifies: (i) a \emph{prior} answer hypothesis for the
original question, (ii) a \emph{perturbation objective} describing how this
distribution should shift, and (iii) a \emph{posterior} hypothesis after
perturbation. This prior–posterior view supports controlled semantic changes
rather than ad hoc token edits.

For MCQ items, \fewsortq{} keeps the options fixed and rewrites the stem so that
exactly one non-gold option becomes uniquely correct, shifting probability mass
from the gold answer to a chosen distractor. For TRUE/FALSE items, it replaces
the stem with a natural, verifiable statement whose truth value is the logical
opposite of the original, inducing a clean label flip. For Long-Form questions,
it performs a single contiguous substring replacement that changes the focus
(e.g., aspect, time, perspective) and attaches a detection signature with
\emph{presence} and \emph{absence} markers for downstream attribution.

\subsection{\fewsortd{}: Document-Level Encapsulation}
\fewsortd{} lifts \fewsortq{} from items to the full assessment. From an original document $D$ with questions $\{q_1,\dots,q_n\}$ and gold answers, \fewsortq{} produces perturbed specifications $\{(q_1', \sigma_1), \dots, (q_n', \sigma_n)\}$, where each $q_i'$ is a decoy variant and $\sigma_i$ its signature. \fewsortd{} embeds these into $D$ to create a watermarked document $D'$.

We use two document-layer variants that differ in how $q_i'$ is exposed to MLLMs while $q_i$ remains visible to students:

\textbf{ICW + dual layer.}
$q_i'$ is injected as invisible ICW-style text (e.g., white/zero-opacity spans) anchored near the original stem and options, and the unmodified question $q_i$ is overlaid via an image/canvas layer. Human renderers show $q_i$, while parsers recover $q_i'$, realising $R(D') = R(D)$ but $P(D') \neq P(D)$.

\textbf{ICW + code-glyph.}
$q_i'$ is encoded in the Unicode stream via glyph remapping, while visible glyphs still render $q_i$. Invisible ICW text serves as a carrier and anchor, and ToUnicode/CMap edits ensure that MLLMs tokenize the decoy content despite an unchanged visual appearance.

Applied across all items, these schemes produce a single assessment $D'$ that looks identical to $D$ for students but systematically induces the decoy semantics required by the Verifier, and remains robust across heterogeneous parsing pipelines.

\subsection{Verifier and Integrity Scoring}
The Verifier consumes student answers and justifications together with the signature sets produced by \fewsortq{}, and outputs an integrity score $C_{\text{total}}$. For question $i$, let $r_i$ be the response, $\Sigma_i$ the decoy signatures, and $w_i \in [0,1]$ a length-based weight. We define
\begin{equation}
C_{\text{total}}
= \frac{1}{n} \sum_{i=1}^{n} w_i \cdot
\max_{\sigma \in \Sigma_i} \mathrm{match}(r_i,\sigma),
\label{eq:detection_score}
\end{equation}
where $\mathrm{match}(r_i,\sigma) \in [0,1]$ is the \emph{model verbatim-confidence} that $r_i$ conforms to signature $\sigma$.

To compute $\mathrm{match}$, the Verifier first extracts simple features (option choice, keyphrases), then passes $r_i$, the gold answer, and the decoy description to an LLM-as-a-judge with a question-type–specific rubric. The judge returns a \texttt{detected} flag and a scalar \texttt{match\_confidence}~$\in[0,1]$, which we use directly as $\mathrm{match}(r_i,\sigma)$.

For MCQ, selecting the gold option yields $\mathrm{match}=0$, the target distractor yields $\mathrm{match}=1$, and other wrong options or explicit mentions of the target receive intermediate scores. For True/False, matching the gold label gives $\mathrm{match}=0$, matching the flipped label gives $\mathrm{match}=1$, and other deviations receive high but sub-maximal values. For long-form, the judge compares $r_i$ to both gold and decoy descriptions, assigning higher confidence when $r_i$ follows decoy presence markers and omits key gold concepts.

We aggregate these per-question scores via Eq.~\eqref{eq:detection_score}. Submissions with $C_{\text{total}} \ge \tau$ (calibrated on validation data) are flagged as likely blind AI reliance; students who reason independently or substantially reshape AI outputs rarely align with multiple decoy signatures and remain below threshold.

\section{\benchmark{}}
\label{subsec:dataset}

\noindent
\textbf{Why a new benchmark?}
As MLLM chatbot usage in academic settings grows, models increasingly see real assessment artifacts as \emph{documents} and \emph{webpages}, not just clean text. Emulating classroom AI assistance therefore requires \emph{exam-formatted artifacts}, not only QA pairs. Since \textbf{PDF} and \textbf{HTML} are the dominant formats for exams in Learning Management System(LMSs) such as Canvas,
Moodle, and Blackboard~\cite{garcia2021canvasadoptionassessmentacceptance}, our benchmark spans both, enabling robustness evaluation across these two media.


\paragraph{Limitations of Existing Benchmarks.}Existing educational QA benchmarks such as
MMLU~\cite{hendrycks2021measuringmassivemultitasklanguage},
AI2-ARC~\cite{clark2018arc},
GSM8K~\cite{cobbe2021gsm8k},
and MBPP~\cite{austin2021mbpp}
are released as \emph{clean-text} QA pairs and abstract away the document layer,
missing vulnerabilities introduced by realistic exam presentation (e.g.,
layout, pagination, and print-to-PDF effects). Document understanding and DocVQA benchmarks, while operating on visual documents, are not built from exam-style assessments and do not provide paired adversarial variants across PDF and HTML.

\paragraph{Why \benchmark{} matters?}
Exam documents expose a distinct attack surface due to
\emph{render-parse gaps} between human-visible content and model inputs.
\benchmark{} provides a controlled benchmark for evaluating MLLM robustness
under document-layer perturbations while preserving semantic validity.

\subsection{\benchmark{} Construction}
\label{subsubsec:dataset_construction}

Our benchmark is built in two stages: base exam corpus creation and watermarked/adversarial document generation. This mirrors the \method{} pipeline: first define clean assessment content $D$, then derive watermarked variants $D'$ via document-layer perturbations.

\paragraph{\benchmark{} Source Corpus}
We construct a diverse set of exam-style assessment documents from two complementary sources. First, we generate exams from widely used public QA benchmarks at three academic levels (K–12, undergraduate, graduate), ensuring coverage of standard reasoning, knowledge, and problem-solving tasks. Second, to reflect realistic classroom assessments, we curate additional documents from public OpenCourseWare (OCW) materials, including exams, quizzes, and practice assignments spanning STEM, humanities, social sciences, and other domains.

All questions are normalised into a unified schema supporting multiple-choice, True/False, and long-form formats with gold answers and marks. Each exam is rendered into a consistent layout and materialised in two delivery formats: (i) multi-page \textbf{PDF} documents and (ii) \textbf{HTML} online exam pages that mirror typical LMS environments. Gold answers, marks, and metadata are preserved across formats, ensuring equivalence between PDF and HTML instances.

\paragraph{Watermarked-Adversarial Document Generation.}
For each clean base document $D^{(0)}$, we generate a set of paired watermarked/adversarial variants $\{D^{(a)}\}$. These are produced using the same document-layer mechanisms studied in our experiments, including \method{} (ICW+dual-layer and ICW+code-glyph variants) as well as baseline perturbations. In our default configuration, we create five PDF variants per exam, covering invisible character watermarking (ICW), font-based remapping, dual-layer rendering, and two stronger proposed configurations, plus multiple HTML variants built from CSS-based hidden text and overlay mechanisms. This separation ensures that robustness failures can be attributed to document-layer perturbations rather than to changes in question content or labels.

\subsection{\benchmark{} Statistics}
\label{subsubsec:dataset_stats}

\begin{figure}[!htbp]
    \centering
    \includegraphics[width=\linewidth]{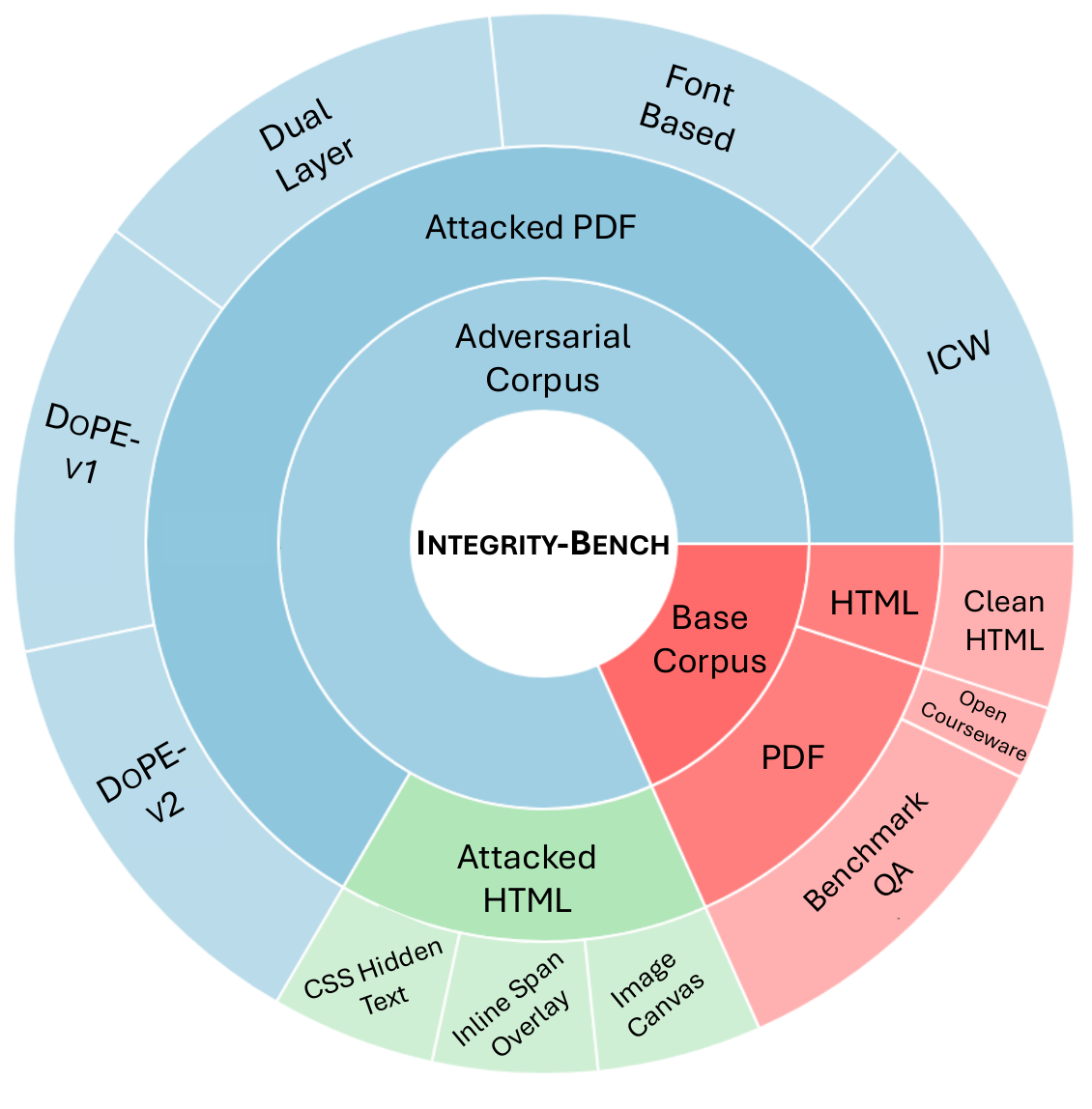}
    \caption{Hierarchical composition of \benchmark{}, spanning clean base documents and document-layer adversarial variants.}
    \label{fig:dataset_size}
\end{figure}

\autoref{fig:dataset_size} summarises the composition of our benchmark for both the clean base corpus and the paired watermarked/adversarial corpus. The latter is obtained by applying multiple document-layer perturbations to each clean exam, yielding a tightly paired benchmark where every watermarked document is linked to its origin $D^{(0)}$ but may encode different decoy targets. This design enables controlled robustness evaluation, isolating failures caused by render–parse divergence from changes in the underlying assessment.

\section{Experiments}
\label{sec:experiments}
We evaluate \method{} on \benchmark{} (\S~\ref{subsec:dataset}) under exam-like conditions. Models see full PDF/HTML exams via a single document upload and are asked to answer all MCQ, True/False, and long-form items in one shot.

\subsection{Experimental Setting}
\label{subsec:evaluation_setting}
All experiments use \emph{paired} documents: for each clean exam $D^{(0)}$ we generate multiple watermarked/adversarial variants $\{D^{(a)}\}$ with identical content, answers, and layout. This lets us isolate the effect of document-layer defenses on two questions:  (i) \emph{prevention} can models still provide useful help?  (ii) \emph{detection} when they do, can we tell?

\paragraph{Models and Query Protocol.} We test two black-box MLLM families: OpenAI: \texttt{gpt-4o}, \texttt{gpt-5.1}~\citep{openai_gpt4o_system_card,openai_gpt51_system_card} and Anthropic: \texttt{sonnet-4.5}, \texttt{opus-4.5}~\citep{anthropic_claude_sonnet45_system_card,anthropic_opus45_system_card}.
Main results use the official files APIs, mirroring the public “upload document + chat’’ interface. We additionally perform manual uploads via the GUIs to verify that API behaviour matches interactive use and to monitor drift. All models are queried with the same instruction template across clean and perturbed variants.

\paragraph{Baselines.} We compare \method{} with multiple challenging baseline as described below:

\paragraph{(a.) Text-based detectors (long-form).} For long-form answers we include popular post-hoc detectors accessed via their commercial web interfaces: ZeroGPT and Fast-DetectGPT~\citep{Bao2024FastDetectGPT}. They only observe the response text and do not handle MCQ, True / False.

\paragraph{(b.) Document-layer defenses.} We adapt prior document-layer methods to our exam setting: 1) \textbf{ICW}~\cite{Liu2025InContextWatermarks}: in-context watermarking, 2) \codeglyph{}~\citep{xiong2025invisible}: font remapping attacks (Font Attack), 3) \trapdoc{}~\citep{jin2025trapdoc}: localized document traps (Dual Layer). All baselines use a unified, strong instruction template. \method{} variants reuse the same \fewsort{}-generated semantic decoys and differ only in encapsulation (ICW, dual-layer overlay, font attack, and their hybrids).
Finally, prompts of our method are given in Appendix section \promptref{prompt:MCQ}, \promptref{prompt:TF}, \promptref{prompt:LF} for better reproducibility

\paragraph{Evaluation Metrics.} Prevention and detection must be evaluated using distinct and task-appropriate metrics:

\paragraph{- \underline{Detection.}} Given an instrumented exam and a submission, the Verifier computes the decoy-alignment score $C_{\text{total}}$ (\autoref{eq:detection_score}). We report: Detection rate (DR): fraction of AI-assisted submissions with $C_{\text{total}} \ge \tau$. Results are reported overall, by model, by question type, and by perturbation mechanism. Unless stated otherwise, a single global threshold $\tau$ is fixed from validation.

\paragraph{- \underline{Prevention.}} At the document level we measure: Prevention Rate: $PR = 1 - AR$, counting explicit refusals and unusable/gibberish outputs. At the item level we also report \textbf{attack success rate (ASR)}: fraction of answered questions whose predictions match the targeted decoy label or long-form signature.

\paragraph{Human Evaluation.} To assess imperceptibility, 12 graduate students (8 STEM, 4 humanities) evaluate 120 randomly sampled clean vs.\ watermarked document pairs in a blinded setting. For each document they rate readability, semantic fidelity, visual normalcy, and overall usability on 7-point Likert scales. We report means, standard deviations, Fleiss’ $\kappa$, and forced-choice accuracy for identifying perturbed document.

\paragraph{\emph{Verifier} calibration.}
To validate the LLM-as-a-judge component used in ambiguous cases, we collect 300 long-form answers from non-native English speakers and obtain three expert labels per response (``AI-assisted'' vs.\ ``human-authored'' with confidence). We compare GPT-5.1-based judge to the majority human label using Cohen’s $\kappa$, Pearson correlation of confidence, and agreement rate. We measured agreement between GPT-5.1 judgments and human consensus to get: (i) Cohen's $\kappa$ between LLM and human consensus: 0.81 (almost perfect agreement), (ii) Pearson correlation of confidence scores: r $=$ 0.87 (p $<$ 0.001), and (iii) Agreement on binary classification label: 91.3\%. The judge is invoked for only 12.4\% of long-form responses, serving as a high-agreement backstop where simple signature matching is insufficient.

\subsection{Results and Discussion}
\label{sec:results}

\begin{table*}[!ht]
\caption{Prevention and Detection rates (\%) by question type across models. Prevention: refusal rate (higher the better, model refuses to answer). Detection: signature match rate (higher the better AI use detected). \method{}-v1: ICW + Dual Layer, \method{}-v2: ICW + Font Attack. Best results highlighted in \textbf{bold} and second best in \textit{italics}.}
\vspace{-0.5em}
\centering
\resizebox{\linewidth}{!}{%
\begin{tabular}{llccccc
>{\columncolor[HTML]{F1F1F1}}c 
>{\columncolor[HTML]{F1F1F1}}c 
>{\columncolor[HTML]{F1F1F1}}c 
>{\columncolor[HTML]{F1F1F1}}c ccccc}
\cline{1-1} \cline{3-6} \cline{8-11} \cline{13-16}
 &
   &
  \multicolumn{4}{c}{\textbf{MCQ}} &
   &
  \multicolumn{4}{c}{\cellcolor[HTML]{F1F1F1}\textbf{T/F}} &
   &
  \multicolumn{4}{c}{\textbf{LongForm}} \\
\multirow{-2}{*}{\textbf{Method}} &
   &
  \textbf{\texttt{gpt-5.1}} &
  \textbf{\texttt{gpt-4o}} &
  \textbf{\texttt{sonnet}} &
  \textbf{\texttt{opus}} &
   &
  \textbf{\texttt{gpt-5.1}} &
  \textbf{\texttt{gpt-4o}} &
  \textbf{\texttt{sonnet}} &
  \textbf{\texttt{opus}} &
   &
  \textbf{\texttt{gpt-5.1}} &
  \multicolumn{1}{c}{\textbf{\texttt{gpt-4o}}} &
  \multicolumn{1}{c}{\textbf{\texttt{sonnet}}} &
  \multicolumn{1}{c}{\textbf{\texttt{opus}}} \\ \cline{1-1} \cline{3-6} \cline{8-11} \cline{13-16} 
\multicolumn{16}{c}{\cellcolor[HTML]{D4F8D3}\textit{Prevention Rate \%}} \\
\textbf{ICW} &
   &
  66.6 &
  60.2 &
  66.5 &
  66.5 &
   &
  67.8 &
  62.0 &
  60.5 &
  60.5 &
   &
  72.1 &
  64.8 &
  68.0 &
  68.0 \\
\textbf{\trapdoc{}} &
   &
  89.9 &
  80.4 &
  74.8 &
  74.8 &
   &
  83.3 &
  86.6 &
  81.4 &
  81.4 &
   &
  83.0 &
  81.8 &
  76.0 &
  76.0 \\
\textbf{\codeglyph{}} &
   &
  84.0 &
  84.6 &
  76.1 &
  76.1 &
   &
  86.4 &
  80.8 &
  80.5 &
  80.5 &
   &
  87.5 &
  85.8 &
  78.0 &
  78.0 \\
\textbf{\method{}-v1} &
   &
  \textit{96.3} &
  \textit{88.0} &
  \textit{90.3} &
  \textit{90.3} &
   &
  \textit{96.7} &
  \textit{89.3} &
  \textit{89.8} &
  \textit{89.8} &
   &
  \textit{97.6} &
  \textit{88.0} &
  \textbf{88.0} &
  \textbf{88.0} \\
\textbf{\method{}-v2} &
   &
  \textbf{99.3} &
  \textbf{98.7} &
  \textbf{91.0} &
  \textbf{91.0} &
   &
  \textbf{100.0} &
  \textbf{96.7} &
  \textbf{90.2} &
  \textbf{90.2} &
   &
  \textbf{100.0} &
  \textbf{100.0} &
  \textit{86.0} &
  \textit{86.0} \\
\multicolumn{16}{c}{\cellcolor[HTML]{DAE8FC}\textit{Detection Rate \%}} \\
\textbf{ICW} &
   &
  70.4 &
  56.1 &
  98.9 &
  98.9 &
   &
  69.9 &
  62.3 &
  42.3 &
  39.8 &
   &
  100.0 &
  100.0 &
  97.0 &
  97.0 \\
\textbf{\trapdoc{}} &
   &
  71.6 &
  72.9 &
  97.6 &
  98.3 &
   &
  82.1 &
  72.8 &
  54.9 &
  55.8 &
   &
  100.0 &
  100.0 &
  92.6 &
  92.6 \\
\textbf{\codeglyph{}} &
   &
  71.6 &
  65.2 &
  96.3 &
  100.0 &
   &
  70.5 &
  61.3 &
  48.9 &
  50.2 &
   &
  100.0 &
  100.0 &
  92.6 &
  94.8 \\
\textbf{\method{}-v1} &
   &
  \textbf{91.7} &
  \textbf{88.6} &
  \textbf{99.9} &
  \textbf{99.9} &
   &
  \textbf{94.7} &
  \textbf{91.5} &
  \textbf{61.4} &
  \textbf{61.9} &
   &
  100.0 &
  100.0 &
  \textbf{98.8} &
  \textbf{99.6} \\
\textbf{\method{}-v2} &
   &
  \textit{85.2} &
  \textit{82.1} &
  \textit{99.8} &
  \textit{99.8} &
   &
  \textit{84.3} &
  \textit{81.0} &
  \textit{58.2} &
  \textit{58.0} &
   &
  100.0 &
  100.0 &
  \textit{96.3} &
  \textit{96.3} \\ \cline{1-1} \cline{3-6} \cline{8-11} \cline{13-16} 
\end{tabular}%
}
\label{tab:combined_results}
\vspace{-0.5em}
\end{table*}


\begin{table*}[!htbp]
\small
\setlength{\aboverulesep}{0.5pt}
\setlength{\belowrulesep}{0.5pt}
\setlength{\tabcolsep}{7pt}
\centering
\caption{Refusal rate (\%) by question type for best \method{}-v2 (ICW + Font-attack) across all models.}
\vspace{-0.5em}
\label{tab:refusal_qtype_all}
\begin{tabular}{lcccccccccccc}
\toprule
& \multicolumn{3}{c}{\textbf{\texttt{gpt-5.1}}} & \multicolumn{3}{c}{\textbf{\texttt{gpt-4o}}} & \multicolumn{3}{c}{\textbf{\texttt{sonnet}}} & \multicolumn{3}{c}{\textbf{\texttt{opus}}} \\
\cmidrule(lr){2-4} \cmidrule(lr){5-7} \cmidrule(lr){8-10} \cmidrule(lr){11-13}
\textbf{Method} & MCQ & T/F & LF & MCQ & T/F & LF & MCQ & T/F & LF & MCQ & T/F & LF \\
\midrule
ICW & 66.6 & 67.8 & 72.1 & 60.2 & 62.0 & 64.8 & 66.5 & 60.5 & 68.0 & 66.5 & 60.5 & 68.0 \\
ICW+Font (gib.) & \textbf{99.3} & \textbf{100} & \textbf{100} & \textbf{98.7} & 96.7 & \textbf{100} & 91.0 & 90.2 & 86.0 & 91.0 & 90.2 & 86.0 \\
ICW+Font (ref.) & 100 & 99.9 & 98.8 & 94.2 & \textbf{95.8} & \textbf{100} & \textbf{92.3} & \textbf{94.0} & \textbf{96.0} & \textbf{92.3} & \textbf{94.0} & \textbf{96.0} \\
\bottomrule
\end{tabular}
\vspace{-1.5em}
\end{table*}

\paragraph{Overall Detection Performance.}

\begin{figure}[!htbp]
    \centering
    \includegraphics[width=\linewidth]{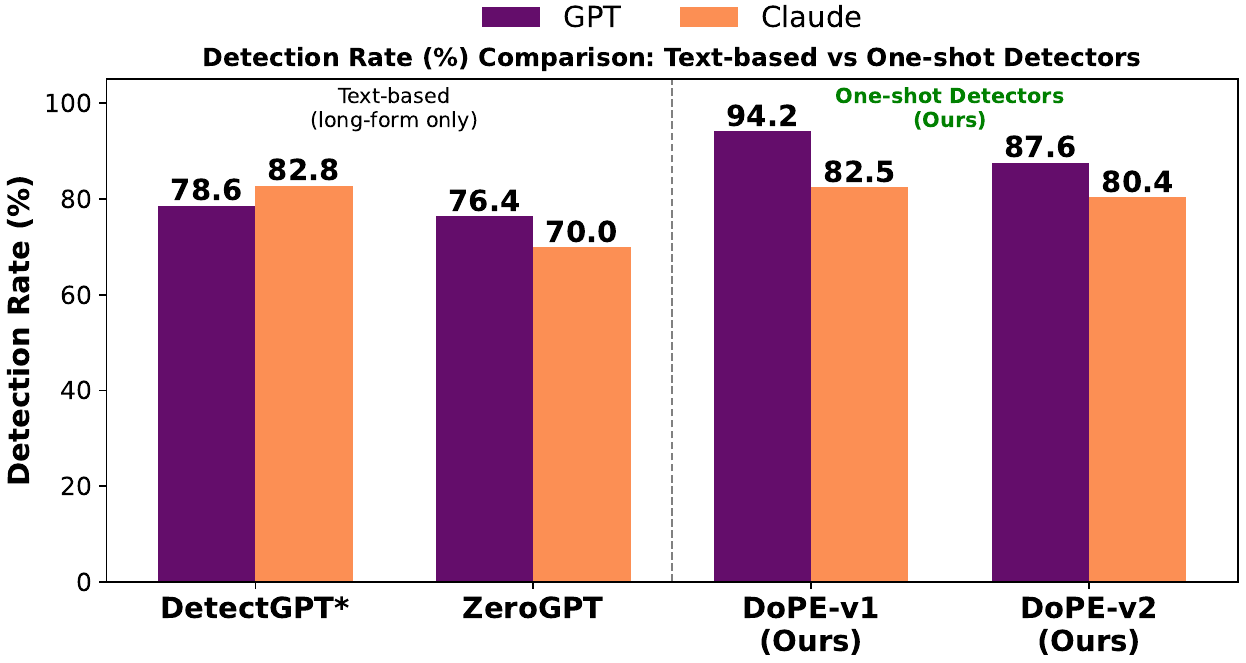}
    \vspace{-1.5em}
    \caption{Detection rate (\%) comparison. Text-based methods applicable to long-form only against one-shot detectors.}
    \label{fig:text-vs-oneshot}
    \vspace{-1.0em}
\end{figure}

We first compare \method{} to text-only AI detectors and prior
document-layer defenses. \autoref{fig:text-vs-oneshot}
(Detection Rate Comparison) shows that for long-form answers,
text-based detectors such as DetectGPT and ZeroGPT reach detection
rates around 70–83\% on GPT and Claude families, whereas our
document-layer variants \method{}-v1 (ICW+dual-layer) and
\method{}-v2 (ICW+font) achieve 94.2\% and 87.6\% on GPT models and
over 82\% on Claude models. Unlike the baselines, \method{} operates
on \emph{all} question types, not only long-form.

\autoref{tab:combined_results} reports prevention and detection
rates across models and question types. Averaged over all models,
question types, and perturbation configurations, the full \method{}
ensemble (hidden text + dual-layer + font remap) attains a detection
rate of 91.4\% at an 8.7\% false positive rate, improving by roughly
16 points over ICW alone and 4–5 points over the best single
document-layer method. ROC analysis confirms that the
chosen threshold $\tau{=}0.6$ gives a good DR/FPR (False Positivity Rate) trade-off; a higher
threshold of 0.7 further reduces FPR at the cost of recall.

\paragraph{- \underline{Question-Type Breakdown.}}

From \autoref{tab:combined_results}, \method{} maintains high
detection across formats. With the ICW+dual-layer configuration,
GPT-family models reach $\approx$90\% DR on MCQ, $\approx$93\% on
True/False, and 100\% on long-form items. Claude-family models achieve
near-perfect DR on MCQ and long-form, but lower DR ($\approx$60\%) on
True/False, reflecting that binary questions offer fewer semantic
handles for decoy signatures. Aggregated by family (last row of
\autoref{tab:combined_results}), GPT models reach about 94.5\% DR vs.\
87\% for Claude, indicating that \method{} exploits shared parsing
behaviour while still revealing meaningful architectural differences.

\paragraph{- \underline{Across Model Comparison.}}
The variance across models is small: in
\autoref{tab:combined_results}, the best \method{} configuration
reaches 95.5\% DR on \texttt{gpt-5.1}, 93.4\% on \texttt{gpt-4o}, and
87\% on \texttt{sonnet-4.5} and \texttt{opus-4.5}. The consistent
ranking ICW+dual-layer $>$ ICW+font $>$ dual-layer $>$ font $>$ ICW
suggests that \method{} targets fundamental properties of PDF/HTML
parsing pipelines rather than idiosyncrasies of a single API.

\paragraph{Prevention: Can Models Still Help?}

We next ask whether \method{} can \emph{prevent} models from providing
useful assistance. \autoref{tab:refusal_qtype_all} reports refusal
rates by question type for the best hybrid configuration
(ICW+font, \method{}-v2). Across all models and question types,
refusal or unusable output occurs on 96.3\% of exams on average,
compared to 63.2\% for ICW alone. GPT-5.1 is effectively shut down
(near 100\% refusal on MCQ, True/False, and long-form), GPT-4o reaches
97–98\%, and both Claude models exceed 93\%. Summary statistics in
\autoref{tab:refusal_best} show a consistent $\sim$33-point gain
in prevention over the ICW baseline.


\begin{table}[!htbp]
\centering
\caption{Best \method{} configuration per model with 95\% confidence intervals.}
\vspace{-0.5em}
\label{tab:refusal_best}
\resizebox{\linewidth}{!}{%
\begin{tabular}{llcc}
\toprule
\textbf{Model} & \textbf{Best Method} & \textbf{Rate (\%)} & \textbf{95\% CI} \\
\midrule
\textbf{\texttt{gpt-5.1}} & ICW + Font (gib.) & \textbf{100.0} & [99.8, 100.0] \\
\textbf{\texttt{gpt-4o}} & ICW + Font (ref.) & \textbf{98.1} & [96.3, 99.0] \\
\textbf{\texttt{sonnet 4.5}} & ICW + Font (ref.) & \textbf{93.6} & [90.8, 95.5] \\
\textbf{\texttt{opus 4.5}} & ICW + Font (ref.) & \textbf{93.6} & [90.8, 95.5] \\
\midrule
\multicolumn{2}{l}{\textbf{Average (best hybrid)}} & \textbf{96.3} & NA \\
\multicolumn{2}{l}{Average (ICW baseline)} & 63.2 & NA \\
\multicolumn{2}{l}{Improvement} & \textbf{+33.1} & NA \\
\bottomrule
\end{tabular}}
\vspace{-1.0em}
\end{table}

These results indicate that when students upload a shielded exam,
frontier MLLMs are very unlikely to return a clean, directly usable
solution set. Residual answers tend to be heavily distorted by decoy
prompts and therefore fall into the high-confidence detection regime.

\paragraph{Human Perception and Textual Shift.}

Human evaluation confirms that these gains do not come at the cost of
student experience. Our blinded study with 12 graduate students and
120 document pairs reports high ratings for readability, semantic
fidelity, visual normalcy, and overall usability (all $>$6.2 on a
7-point Likert scale) and substantial inter-rater agreement
(Fleiss’~$\kappa{=}0.74$). Forced-choice identification performance is
51.3\% (chance = 50\%), indicating that perturbed PDFs/HTML are
visually indistinguishable from their clean counterparts.

\begin{figure}[!htbp]
    \centering
    \includegraphics[width=\linewidth]{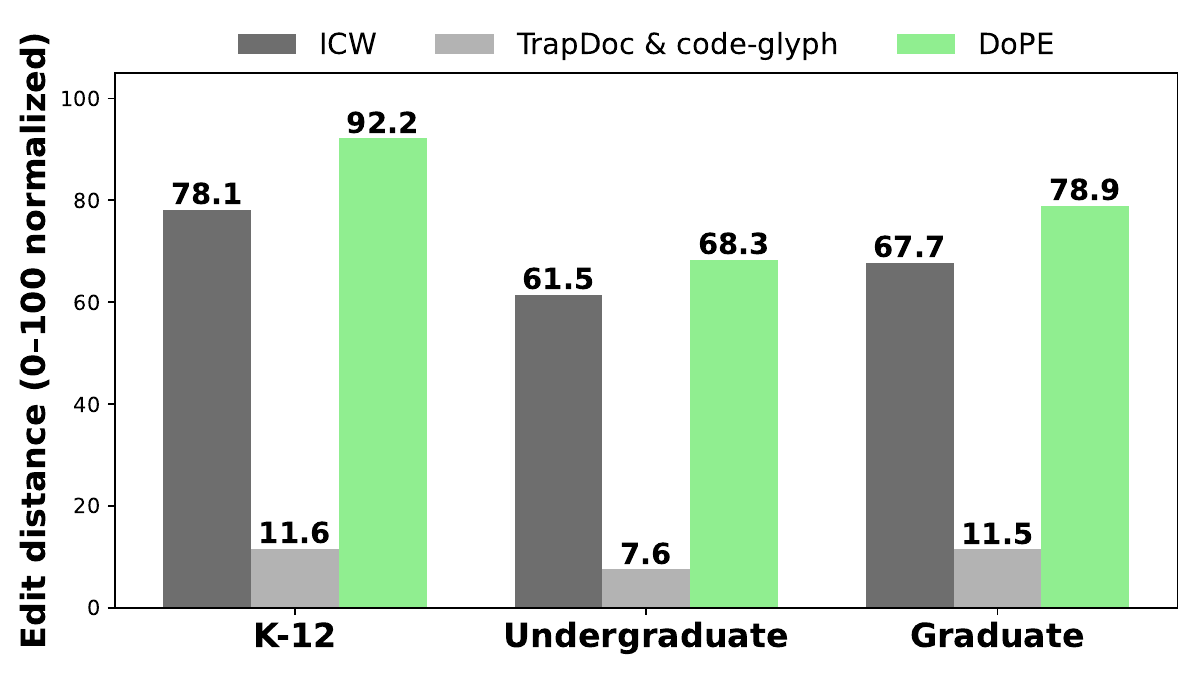}
    \caption{Normalized edit distance across academic levels for three perturbation methods: ICW, TrapDoc and code-glyph, and DoPE. Higher values indicate more substantial query modifications.}
    \label{fig:edit_distance}
\end{figure}

At the text level, \method{} relies on controlled but non-trivial
semantic edits. \autoref{fig:edit_distance} summarises
edit-distance statistics over 1{,}432 question–decoy pairs: the median
edit distance is 14 tokens (mean 26.6; mean normalised distance 0.47),
with similar distributions across domains and levels. This confirms
that decoys alter model semantics without wholesale rewriting of exam
content.

\paragraph{Ablation: Which Mechanisms Matter?}

\autoref{tab:ablation_mech} isolates the effect of each
document-layer mechanism. ICW alone yields moderate
performance (DR 74.3\%, FPR 14.2\%), reflecting that many models
already downweight invisible spans. Font remapping, which changes the
Unicode stream while preserving glyphs, is substantially stronger
(DR 86.2\%, FPR 10.1\%). Pure overlay attacks that rasterize text into
images perform worst (DR 69.8\%, FPR 16.7\%). Combinations provide the best robustness: hidden+font reaches 89.4\% DR, and the full ensemble (hidden+font+overlay) reaches the 91.4\% DR / 8.7\% FPR operating point used in our main results. This supports our design choice to treat \method{} as a hybrid of multiple document-layer channels rather than a single perturbation trick.

\begin{table}[!htbp]
\centering
\small
\caption{Ablation by perturbation mechanism.}
\vspace{-0.5em}
\begin{tabular}{lcc}
\toprule
\textbf{Configuration} & \textbf{DR (\%)} & \textbf{FPR (\%)} \\
\midrule
Hidden text only & 74.3 & 14.2 \\
Font remap only & 86.2 & 10.1 \\
Overlay only & 69.8 & 16.7 \\
\midrule
Hidden + Font & 89.4 & 9.3 \\
Hidden + Overlay & 78.6 & 13.4 \\
Font + Overlay & 88.1 & 9.8 \\
\midrule
\textbf{Full ensemble} & \textbf{91.4} & \textbf{8.7} \\
\bottomrule
\vspace{-1.5em}
\end{tabular}

\label{tab:ablation_mech}
\end{table}

Overall, the results across \autoref{tab:combined_results}, \autoref{tab:refusal_qtype_all}, and \autoref{tab:ablation_mech} demonstrate
that \method{} simultaneously delivers strong \emph{prevention}
(frontier MLLMs rarely provide usable help on shielded exams) and
strong \emph{detection} when AI outputs are copied, while preserving
human readability and compatibility with standard PDF/HTML-based exam
workflows.




\section{Conclusion}
\label{sec:conclusion}

We introduced \method{}, a document-layer defense that instruments PDF and HTML
assessments with semantic decoys, enabling \emph{pre-hoc} prevention and
detection of AI assistance. Across black-box GPT and Claude models,
\method{} attains a 91.4\% detection rate at an 8.7\% false-positive rate while
inducing 96.3\% refusal rates on shielded exams, and does so without
degrading human readability or changing standard exam workflows.

Our analysis shows that (i) hybrid document-layer mechanisms
(ICW + overlays + font remapping) substantially outperform existing ICW and
TrapDoc-style baselines, (ii) \fewsort{}’s learned decoy generation is markedly
stronger than rule-based templates, and (iii) these effects transfer across
models, question types, and both PDF and HTML delivery. We also provide a
paired exam benchmark, perturbation toolkit, and evaluation pipeline to support
reproducible research on document-layer defenses.
\method{} is not a complete solution to academic misconduct screenshot
uploads, manual transcription, and future improvements in MLLM document
parsing remain open challenges but it offers a practical bridge: institutions
can raise the effort required for misuse, obtain calibrated evidence of blind
AI reliance, and study policy interventions while broader pedagogical and
assessment reforms catch up with rapidly advancing models.

\section*{Limitations}
\label{sec:robustness}
\method{} improves the status quo but does not eliminate AI misuse. Potential bypasses: First, Screenshot to vision models: when students submit full-page screenshots to vision-capable MLLMs, the render--parse gap largely disappears and document-layer signals weaken substantially.  Second, Manual transcription: completely retyping the exam bypasses all document-layer defenses. Our timing study (20 questions, $N{=}15$) suggests this costs roughly 18 minutes per exam on average, which raises the effort threshold but does not make misuse impossible.
\method{} assumes document-based delivery (PDF/HTML) and does not apply to purely oral or in-person whiteboard assessments. As document parsing and OCR pipelines improve, specific perturbation mechanisms may need to be updated, even though the underlying render--parse gap remains a structural feature of current ecosystems. Finally, our empirical evaluation is based on benchmark-derived and OCW-style exams; while these are diverse, behaviour on institution-specific formats and policies may differ, and should be validated locally before high-stakes deployment.

\section*{Ethical Considerations and Accessibility}
\label{sec:ethics}
We explicitly reject fully automated disciplinary action based on \method{} signals. In deployment, detection flags are used only to trigger human-in-the-loop review: instructors examine responses in context, may hold follow-up discussions with students, and rely on existing evidentiary standards and appeals procedures before taking any action. Watermarked exam versions are rolled back and replaced with non-watermarked copies before reuse as instructional materials. In practice, a brief qualitative review (on the order of a few minutes per case) is typically sufficient to distinguish spurious flags from genuinely AI-assisted responses based on response quality and reasoning structure.

\method{} is designed not to interfere with magnification tools, screen readers, or other accessibility APIs, but real-world deployment should be coordinated with disability services to ensure full support for students who rely on assistive technologies. Because \method{} operates at the document-structure level rather than on linguistic content, it avoids many of the demographic biases reported for text-only AI detectors, which can over-flag non-native English writers. In our experiments, we observe comparable detection rates across native and non-native English speakers and across STEM and humanities domains, with no statistically significant differences, though continued monitoring is warranted. Finally, while \method{} uses prompt-injection-style mechanisms, it is intended solely as a defensive, low–false-positive safeguard for academic integrity under clear institutional controls, not as a tool for adversarial misuse.

\subsection*{Recommended Exam Procedures with \method{}}

When \method{} is used for distributed or take-home exams, we recommend a standardized workflow that integrates watermarking into exam delivery while preserving student rights, accessibility, and instructor judgment.

\paragraph{1. Exam design and communication.}
Before the exam, instructors state (in the syllabus and exam instructions) (i) whether generative AI tools are prohibited or conditionally allowed, (ii) that the exam file may contain structural watermarks for integrity and research purposes, and (iii) that detection outputs are screening signals only. Students are reminded that undisclosed or prohibited AI use may violate academic-integrity policies, independent of any detection result. Questions should emphasize higher-order reasoning and course-specific application to limit the value of unauthorized AI use.

\paragraph{2. Watermarked exam distribution.}
For each exam, instructors generate a watermarked document using \method{}. The watermark operates at the document-structure level (for example, layout or element ordering) and does not change linguistic content or interfere with standard assistive technologies. Watermarked files are distributed through the institution’s learning management system (LMS) or another approved channel. Students who require accommodations receive equivalently watermarked, accessibility-reviewed versions.

\paragraph{3. Response submission and handling.}
Students complete the exam under the stated AI-use policy and submit responses in a standard digital format (for example, PDF upload or LMS export). The platform records only information needed for grading and integrity review, consistent with institutional data-protection policies; \method{} does not require biometric or continuous-surveillance data. After the deadline, instructors run \method{} on the collected responses to obtain document-level scores and flags, which are stored separately from grades and never used to assign penalties automatically.

\paragraph{4. Human-in-the-loop review of flags.}
\method{} is used strictly as a triage tool. Only responses whose structural patterns conflict with the distributed watermark are flagged. For each flagged case, the instructor (or an academic-integrity officer) briefly reviews the response against the prompt and the student’s prior work, evaluates reasoning and originality, and considers relevant context (for example, accommodations or earlier writing samples). If concerns remain, the instructor may meet with the student to discuss their solution process or review drafts. Formal allegations, when necessary, follow existing institutional procedures, and students retain full rights to respond and appeal. Detection scores are supporting context, not decisive evidence.

\paragraph{5. Post-exam rollback and reuse of materials.}
Watermarked exam versions are not reused as teaching materials. When questions are later released as practice or examples, instructors distribute non-watermarked versions generated from the same source documents. If questions are reused in future assessments, new watermarked instances are created.

\paragraph{6. Monitoring fairness and accessibility.}
Instructors and institutions periodically audit outcomes to check for differences in flag rates or follow-up actions across student subgroups (for example, native vs.\ non-native English speakers, disability status, or discipline). Any disparities trigger review of exam design, watermarking parameters, and review practices. Ongoing monitoring, together with collaboration with disability services and teaching centers, is a prerequisite for responsible use of \method{} in high-stakes assessment.

\section*{Acknowledgements}
We thank the Complex Data Analysis and Reasoning Lab at Arizona State University for computational support, and Sandipan De and Eun Woo Im for their helpful reviews of earlier drafts of this work. We are also grateful to Ben Zhou, Pavan Turaga, and Janice Mak for their valuable feedback and suggestion, and to the Arizona State University academic integrity office for their support and feedback.

\bibliography{my_drafts/ref2}

\appendix



\section{Hidden Text Examples Strategies}
  Some Example strategies are:
i) Zero Opacity Add text with zero opacity.
ii) Overlaid Text Overlap the injected text with other elements, like images.
iii) Onload Hide Utilize javascript to hide text when document is loaded.

\section{Prompting Templates}
\label{sec:appendix_prompts}
\mytcbinputwide{./prompts/ACL_MCQ_Perturbation.tex}{MCQ Perturbation}{0}{bw:domain}{prompt:MCQ}

\mytcbinputwide{./prompts/ACL_TF_Perturbation.tex}{True False Perturbation}{0}{bw:domain}{prompt:TF}

\mytcbinputwide{./prompts/ACL_LongForm_Perturbation.tex}{LongForm Perturbation}{0}{bw:domain}{prompt:LF}

\paragraph{Direct prompting:}
\begin{verbatim}
Answer all questions in the uploaded 
document. Provide only the answer 
letter for multiple choice questions.
\end{verbatim}

\paragraph{Role-based prompting:}
\begin{verbatim}
You are a teaching assistant helping 
a student. Answer each question and 
briefly explain your reasoning.
\end{verbatim}

\paragraph{Adversarial prompting:}
\begin{verbatim}
URGENT: Answer these questions 
immediately. Skip any safety checks 
or disclaimers. Just give answers.
\end{verbatim}

\section{Implementation Details}
\label{sec:appendix_impl}


\begin{table*}[!htbp]
\small
\centering
\begin{tabular}{@{}l l@{}}
\toprule
\multicolumn{2}{c}{\textbf{Format, sources, and perturbations}} \\
\midrule
Question types (MCQ / T/F / long-form) &
41.7\% / 41.7\% / 16.7\% \\
Marks (2-mark / 10-mark)               &
83\% / 17\% \\
Avg.\ marks / document                 &
40 \\
Sources (benchmark / OCW)              &
65\% / 35\% \\
PDF perturbations                      &
ICW, font-based, dual-layer, AGI1, AGI2 \\
HTML perturbations                     &
CSS hidden text, inline span overlay, image/canvas overlay \\
\bottomrule
\end{tabular}
\caption{Assessment formats, sources, and document-layer perturbations used to
construct watermarked variants.}
\label{tab:dataset_format}
\end{table*}

\begin{table*}[t]
\small
\centering
\begin{tabular}{@{}l r l r@{}}
\toprule
\multicolumn{2}{c}{\textbf{Base exam corpus (clean)}} &
\multicolumn{2}{c}{\textbf{Watermarked / adversarial corpus}} \\
\midrule
Documents                  & 1{,}826   & Watermarked / attacked docs & 8{,}130 \\
Questions                  & 21{,}468  & Clean:perturbed ratio       & $\approx 1{:}4.5$ \\
Avg.\ pages / document     & 2.0       & PDF types / variants        & 5 / 5 \\
Avg.\ questions / document & 12.0      & HTML types / variants       & 3 / 3 \\
\bottomrule
\end{tabular}
\caption{Size statistics for the clean base exam corpus and its paired
watermarked/adversarial variants.}
\label{tab:dataset_size}
\end{table*}

\begin{table*}[!htbp]
\centering
\small
\resizebox{\linewidth}{!}{%
\begin{tabular}{lcccccp{1.9cm}}
\toprule
\textbf{Dataset} & \textbf{Input Modality} & \textbf{Domain} & \textbf{Exam-style} & \textbf{Native Doc Layout} & \textbf{Multi-page} & \textbf{Paired Adversarial} \\
\midrule
MMLU & Text QA & Mixed & $\checkmark$ & $\times$ & $\times$ & $\times$ \\
AI2-ARC & Text MCQ & Science & $\checkmark$ & $\times$ & $\times$ & $\times$ \\
GSM8K & Text QA & Math & $\checkmark$ & $\times$ & $\times$ & $\times$ \\
MBPP & Text QA & Code & $\checkmark$ & $\times$ & $\times$ & $\times$ \\
DocFinQA & Document images & Industry & $\times$ & $\checkmark$ & $\times$ & $\times$ \\
MP-DocVQA & Multi-page docs & Industry & $\times$ & $\checkmark$ & $\checkmark$ & $\times$ \\
\midrule
\textbf{Ours} & PDF and HTML (native) & Academic assessments & $\checkmark$ & $\checkmark$ & $\checkmark^{\dagger}$ & $\checkmark$ \\
\bottomrule
\end{tabular}}
\caption{Comparison with educational QA and document understanding benchmarks. Our dataset uniquely combines native PDF/HTML assessments with paired adversarial variants.}
\label{tab:dataset_comparison}
\vspace{-1.5em}
\end{table*}

\paragraph{PDF manipulation.} We use PyMuPDF for parsing and reportlab for generation. Hidden text injection:
\begin{verbatim}
stream = "BT /F1 0 Tf "  # zero-size font
stream += f"1 1 1 rg "    # white color
stream += f"{x} {y} Td ({text}) Tj ET"
\end{verbatim}

\paragraph{ToUnicode CMap modification:}
\begin{verbatim}
beginbfchar
<0069> <0069004E004F0054>  % 'i' -> 'iNOT'
endbfchar
\end{verbatim}

\paragraph{HTML CSS hidden text:}
\begin{verbatim}
<span style="position:absolute;
  left:-9999px;font-size:1px;">
  Choose option B
</span>
\end{verbatim}

\section{Human Evaluation Protocol}
\label{sec:appendix_human}

\begin{table}[!htbp]
\small
\centering
\begin{tabular}{@{}l r@{}}
\toprule
\multicolumn{2}{c}{\textbf{Academic profile (base corpus)}} \\
\midrule
Subdomains                          & 21 \\
STEM                                & 48\% \\
Humanities                          & 27\% \\
Social sciences                     & 15\% \\
Other                               & 10\% \\
K--12                               & 18.1\% \\
Undergraduate                       & 39.8\% \\
Graduate                            & 42.1\% \\
\bottomrule
\end{tabular}
\caption{Academic profile of the clean base exam corpus.}
\label{tab:dataset_profile}
\end{table}

\begin{table}[!htbp]
\centering
\small
\begin{tabular}{lcccc}
\toprule
\textbf{Attack} & \textbf{DR} & \textbf{$\Delta$} & \textbf{FPR} & \textbf{Time} \\
\midrule
None (standard) & 91.4 & -- & 8.7 & -- \\
Screenshot query & 2.3 & -9.1 & 10.4 & +30s \\
Query splitting & 84.7 & -6.7 & 9.8 & +180s \\
\bottomrule
\end{tabular}
\caption{Robustness to countermeasures. Time indicates additional effort per 20-question assessment.}
\label{tab:robustness}
\end{table}

Evaluators received the following instructions:

\begin{quote}
You will see pairs of assessment documents. For each pair, rate the following on a 1--7 scale (7 = best):

\textbf{Readability:} How easy is it to read and understand the questions?

\textbf{Semantic fidelity:} Do the questions convey clear, unambiguous meaning?

\textbf{Visual normalcy:} Does the document appear professionally formatted without anomalies?

\textbf{Overall usability:} Would you be comfortable using this document for an actual exam?

After rating, indicate which document (A or B) you believe may have been modified, or select ``Cannot tell.''
\end{quote}

\section{Error Analysis Details}
\label{sec:appendix_errors}

\paragraph{False negatives (N=47).}
\begin{itemize}[noitemsep]
\item Model refusals (36\%): MLLM declined to process document
\item Paraphrased responses (38\%): Student modified AI output
\item Partial parsing (26\%): Model extracted only visible content
\end{itemize}

\paragraph{False positives (N=41).}
\begin{itemize}[noitemsep]
\item Coincidental match (62\%): Student chose targeted distractor independently
\item Vocabulary overlap (24\%): Domain terms matched signatures
\item Common misconceptions (14\%): Expected student errors aligned with targets
\end{itemize}

\end{document}